\documentclass[letterpaper, 10 pt, conference]{ieeeconf}  

\IEEEoverridecommandlockouts                              

\usepackage{graphics} 
\usepackage{amsmath} 
\usepackage{amssymb}  
\usepackage{multicol}
\usepackage[bookmarks=true]{hyperref}
\usepackage{xcolor}
\usepackage{booktabs}
\usepackage{graphicx}
\usepackage{svg}
\usepackage{subcaption}
\usepackage{xspace}

\usepackage[backend=bibtex,style=ieee,natbib=true,mincitenames=1,maxcitenames=2]{biblatex}
\newcommand{\I}{\mathbf{I}}

\newcommand{\R}{\mathbf{R}}
\newcommand{\pos}{\mathbf{p}}
\newcommand{\vel}{\mathbf{v}}

\newcommand{\Lie}[1]{\text{GL}_#1(\mathbb{R})}

\newcommand{\SO}{\mathrm{SO}}
\newcommand{\so}{\mathfrak{so}}
\newcommand{\SE}{\mathrm{SE}}

\newcommand{\SOthree}{\ensuremath{\SO(3)}\xspace}
\newcommand{\sothree}{\ensuremath{\so(3)}\xspace}

\newcommand{\SEthree}{\ensuremath{\SE(3)}\xspace}
\newcommand{\SEtwothree}{\ensuremath{\SE_2(3)}\xspace}

\newcommand{\ie}{\textit{i}.\textit{e}., }
\newcommand{\eg}{\textit{e}.\textit{g}. }

\title{\LARGE \bf
A Group Theoretic Metric for Robot State Estimation Leveraging Chebyshev Interpolation
}

\author{Varun Agrawal$^{1}$, Frank Dellaert$^{1}$
    \thanks{$^{1}$Institute for Robotics and Intelligent Machines and School of Interactive Computing, Georgia Institute of Technology, Atlanta, GA 30332 USA.
        {\tt\small \{varunagrawal,frank.dellaert\}@gatech.edu}}%
}

\newtheorem{theorem}{Theorem}

\begin{document}

\maketitle
\thispagestyle{empty}
\pagestyle{empty}

\begin{abstract}

We propose a new metric for robot state estimation based on the recently introduced $\SEtwothree$ Lie group definition. Our metric is related to prior metrics for SLAM but explicitly takes into account the linear velocity of the state estimate, improving over current pose-based trajectory analysis. This has the benefit of providing a single, quantitative metric to evaluate state estimation algorithms against, while being compatible with existing tools and libraries.
Since ground truth data generally consists of pose data from motion capture systems, we also propose an approach to compute the ground truth linear velocity based on polynomial interpolation. Using Chebyshev interpolation and a pseudospectral parameterization, we can accurately estimate the ground truth linear velocity of the trajectory in an optimal fashion with best approximation error. We demonstrate how this approach performs on multiple robotic platforms where accurate state estimation is vital, and compare it to alternative approaches such as finite differences. The pseudospectral parameterization also provides a means of trajectory data compression as an additional benefit. Experimental results show our method provides a valid and accurate means of comparing state estimation systems, which is also easy to interpret and report.
\end{abstract}

\global\long\def\degree{N}

\section{INTRODUCTION}

State estimation is an integral part of many robotic systems. For feedback-loop based controllers, having access to accurate state estimates allows the controller to generate the appropriate control law at a particular time instance. The state vector used in such controllers generally consists of the rotation \& translation which represents the pose, and the linear and angular velocity, representing the first derivative of the pose. The state vector can be extended to include other elements, such as the biases in an inertial state estimator, but we focus on the general case where we have an inertial sensor to measure the angular velocity, and thus we only need to estimate the pose and linear velocity.

To improve state estimation, it is vital to be able to measure and compare various approaches, and benchmark them accordinaly. Almost all current metrics used for evaluating state estimation algorithms are rooted in  SLAM or Visual Odometry, and don't take into account the linear velocity aspect of the state, a key issue in these metrics. Some examples of such metrics are Relative Pose Error (RPE), Absolute Trajectory Error (ATE)~\cite{Sturm12iros_rgbd_slam}, and Discernible Trajectory Error (DTE)~\cite{Lee23arxiv_dte}. \citet{Zhang18iros} consider the linear velocity in their state definition, however they report final metrics for the rotation, translation and velocity independently instead of as a single comprehensive metric. As such, the state estimator performance evaluation is missing a crucial aspect required by the robot controller.

\begin{figure}
    \captionsetup{font=footnotesize}
    \centering{}
    \includegraphics[width=\columnwidth]{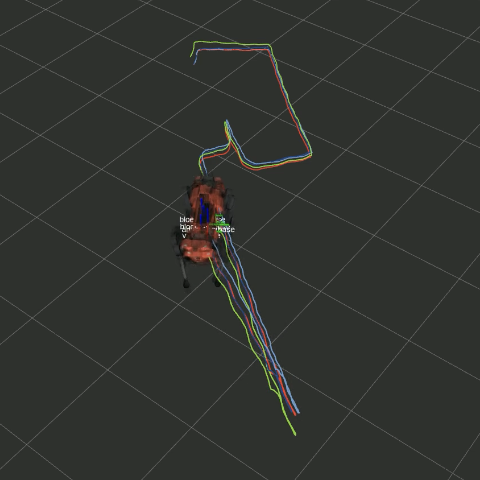}
    \caption{The ability to compare state estimation techniques with a singular value can aid in various design choices of a robot controller, especially those highly dependent on good state estimates, \eg for dynamic walking. Here two state estimate trajectories are visualized for qualitative comparison.}
    \label{fig:estimator-comparison}
\end{figure}

A single metric for evaluation which takes into account all aspects of the state is vital to ensure fair comparison and overall performance. For example, the F1-score used in machine learning and information retrieval~\cite{VanRijsbergen79book} is a comprehensive summary statistic that takes into account both the precision and recall of the classification model. Since the F1-score singularly captures the performance of the model, it is the \textit{de facto} metric reported in information retrieval and supervised classification literature. 

In this work, we propose a singular metric which encapsulates all elements of the generalized state vector. Our metric uses the recently proposed Lie Group $\SEtwothree$\cite{Barrau20icra}, and thus is easy to compute and add to existing software libraries. We refer to this metric as the \textbf{Absolute State Error} (ASE).

Since collecting ground truth data for linear velocity estimation is difficult, we also propose a method for computing the ground truth velocity from the true translation data (generally captured with a motion capture device). While most prior work computes the linear velocity using the method of finite differences~\cite{Wisth19ral_legged_fg}, our proposed method uses differentiation via interpolation. There is a wide choice of polynomials to use as the basis model, such as the Fourier series, B-splines, Bernstein polynomials, or Chebyshev polynomials. In our work, we fit the translation data to Chebyshev polynomials, since they have many of the same benefits of splines~\cite{Park99dissertation}, are known to have optimal performance in a least-square sense, and computing the polynomial derivative is an efficient matrix-vector product~\cite{Hoffman87siam}. To specify that our metric uses a Chebyshev polynomial, we denote this instantiation as the \textbf{Chebyshev Absolute State Error} (C-ASE).

\section{RELATED WORK}

We review the related literature on state estimation, taking into account the error metrics used. All the works report the generalized state vector, with additional components that are problem or platform specific.
Since the related literature on state estimation is quite large, we focus on more recent results in drone and legged robot state estimation.

\citet{Mourikis07icra} is amongst the first state estimation works based on visual-inertial sensor fusion. They report their state estimates as orientation, position and velocity independently without the use of metrics, and without explicitly stating how the true velocity was recorded in hardware experiments.
Similarly, \citet{Lippiello11cdc} use a simulation environment for their experiments and don't report any metrics, rather they show results qualitatively.
\citet{Eckenhoff19ijrr} only report orientation and position Root Mean Square Error (RMSE), and for experiments with velocity they use the Gazebo simulator.
\citet{Svacha19ral} propose an estimator for attitude (orientation) and velocity estimation, each result of which is reported independently with RMSE and Standard Deviation metrics. However, they don't specify how the velocity is computed, other than the data is collected using a motion capture setup.
\citet{Delmerico19icra_uzh_dataset} provide a dataset for high speed autonomous drone racing with only the rotation and position provided. \citet{Svacha20ral} use the platform-provided VIO system to obtain velocity estimates which are used as ground truth substitutes.

For legged robots, \citet{Bloesch12rss_se_imu_leg_kinematics,Bloesch13iros_se_slippery_uneven} used numerical differentiation via finite differences from the motion capture system's translation data.
\citet{Rotella14iros} perform estimation on a humanoid robot with results reported only in simulation.
\citet{Nobili17rss} use a specified velocity of 0.5 m/s, and a custom metric called Drift per Distance Traveled (DDT). DDT considers the translation drift over time, assuming that errors in rotation and velocity will carry over into the translation estimates.
\citet{Hartley18icra_contact_factors} use SimMechanics for evaluating estimation results in simulation. On hardware experiments however, they only report Relative Position Error and show results qualitatively. In their follow up work,~\citet{Hartley18iros_vic} they use the same metric only for translation and use the Cumulative Distribution Function as a measure of translation drift.
\citet{Wisth19ral_legged_fg,Wisth20icra_contact_fg},~\citet{Buchanan21corl,Buchanan23ral},~\citet{Agrawal22humanoids} and \citet{Kim22ral} only report the ATE and RPE, without reporting velocity estimation results.
\citet{Dhedin23icra} run trajectories with specified horizontal linear velocities. They report the RPE for pose estimates, and plot the distribution of the estimated velocities to those of a baseline EKF with access to the motion capture data so show correlations in the estimates.
Similar to~\cite{Wisth19ral_legged_fg,Kim22ral},~\citet{Yang23icra_cerberus} don't report velocity estimates, but whilst claiming to report the Relative \textit{Pose} Error from those papers, they erroneously report the Relative \textit{Translation} Error, which only evaluates the translation estimates. \citet{Fourmy21icra} use a Kalman Filter to compute the base velocity which is used as the baseline, since they are primarily concerned with Center of Mass velocity and inertia estimation.
Probably the most interesting case, \citet{Camurri20frontiers_pronto} use the estimator in a feedback control loop to run walking experiments. Their estimator's state (and particularly velocity) estimates are validated by the performance of the controller. This has the problem of requiring a full locomotion stack, slow iteration times, and difficulty repeating experiments accurately

Polynomial interpolation has seen prior applications in the field of trajectory and state estimation. \citet{Sommer20cvpr} used B-splines to represent trajectories and compute the trajectory derivatives to obtain accurate velocity and acceleration estimates.
\citet{Agrawal21icra_pseudospectral_estimation} used Chebyshev polynomials for representing both the state and the dynamics of a quadrotor in a pseudospectral fashion. They leveraged the Chebyshev differentiation matrix and the quadrotor dynamics model to constrain the valid dynamics states. Similarly,~\citet{Zhu22tsp_chevopt} estimated the coefficients of the Chebyshev polynomial used to represent the trajectory. All these papers perform evaluation in simulation.

From this section, it is apparent that there exists a need for a metric that evaluates all parts of the generalized state vector jointly, as well as an efficient means to compute the linear velocity for hardware experiments.

\section{Preliminaries}

\subsection{Lie Group and Lie Algebra}

A Lie group $\Lie{n}$ is a group which is also a differentiable manifold. It is the set of of all $n \times n$ non-singular real matrices where the group operation is matrix multiplication.

The special orthogonal group in 3D is the Lie group $\SOthree = \{\R \in \Lie{3}~|~\R\R^T = \I, \mathrm{det}\R = +1 \}$. It represents all 3D rotation matrices, with the group operation as matrix multiplication, and the inverse is the matrix transpose. The tangent space at the identity for this manifold is the \textit{Lie algebra} $\sothree$ which is given by the space of all $3\times3$ skew symmetric matrices. The \textit{vee} and \textit{hat} operators are used to convert the matrix to a corresponding vector and vice-versa~\cite{Sola18arxiv}. The \textit{exponential map} associates an element of the Lie algebra to the Lie group $\R$: $\sothree \rightarrow \SOthree$. Conversely, the \textit{logarithm map} converts the Lie group element to the corresponding Lie algebra $\SOthree \rightarrow \sothree$. For more details, we refer the reader to~\cite{Wang08ijrr_lie_group}.

Similarly, the 3D special Euclidean group is given by $\SEthree = \{\mathbf{T} \in \Lie{4}~|~\R \in \SOthree, \pos \in \mathbb{R}^3 \}$ and is the set of rigid transformations on $\mathbb{R}^3$. The group operation, inverse, exponential map and logarithm map are similar to $\SOthree$, with the addition of the translation vector as an additional element of the matrix operations~\cite{Wang08ijrr_lie_group,Sola18arxiv}.

\subsection{Trajectory Evaluation}

Trajectory evaluation first arose as a means to benchmark and evaluate SLAM systems~\cite{Sturm12iros_rgbd_slam}. It is important to not only understand how accurate the system is globally, but also how much it varies in a local sense. For this reason,~\citet{Sturm12iros_rgbd_slam} proposed the Absolute Trajectory Error (ATE) to measure the global consistency, and the Relative Pose Error (RPE) to measure the local accuracy over a given window length.

Given the estimated trajectory $\mathbf{P}_1, ..., \mathbf{P}_n \in \SEthree$ and the true trajectory $\mathbf{Q}_1, ..., \mathbf{Q}_n \in \SEthree$, the Root Mean Square Error (RMSE) for RPE is given by
\begin{align}
    \label{eqn:rpe}
    & \text{RMSE}(\mathbf{E}_{1:N}, \Delta) = \Bigg( \frac{1}{M} \sum_{i=1}^{M} \Vert \textit{trans}(\mathbf{E}_i) \Vert^2 \Bigg)^{1/2}  \\
    & \mathbf{E}_i = (\mathbf{Q}_i^{-1}\mathbf{Q}_{i+\Delta})^{-1}(\mathbf{P}_i^{-1}\mathbf{P}_{i+\Delta})
\end{align}
and for ATE is similarly
\begin{align}
    \label{eqn:ate}
    & \text{RMSE}(\mathbf{F}_{1:N}) = \Bigg( \frac{1}{N} \sum_{i=1}^{N} \Vert \textit{trans}(\mathbf{F}_i) \Vert^2 \Bigg)^{1/2}  \\
    & \mathbf{F}_i = \mathbf{Q}_i^{-1}\mathbf{S}\mathbf{P}_i~~~~~~~~~~~~~~~~~~~~~
\end{align}
where $\textit{trans}$ specifies the translational components of the transform, $\Delta$ is the fixed time interval, $M = N - \Delta$, and $\mathbf{S}$ is the trajectory alignment transform.
\section{STATE ESTIMATION METRIC}

\subsection{State Definition}

To derive our metric, it is important we define our state vector. Different types of robots can have varrying state vector definitions, hence we focus on the general case which is required for robot controllers. We make the assumption that an inertial sensor is available to directly measure the angular velocity of the robot.
For a general robotic system, we define the generalized state vector as
\begin{equation}
    \label{eqn:gen_state}
    \mathbf{x}_i \triangleq [\R_i, \pos_i, \vel_i]
\end{equation}
where $\R_i \in \SOthree$ is the rotation, $\pos_i \in \mathbb{R}^3$ is the translation vector, and $\vel_i \in \mathbb{R}^3$ is the linear velocity at time index $i$.

\subsection{$\SEtwothree$ Group}

Our metric leverages the $\SEtwothree$ Lie group as proposed by~\cite{Barrau20icra}. This group is also termed as ``direct spatial isometries''.
Similar to how $\SEthree$ extends $\SOthree$ with the addition of the translation vector, $\SEtwothree$ analogously extends $\SEthree$ with the linear velocity vector.

We tweak the $\SEtwothree$ group definition in~\cite{Barrau20icra} to be equivalently defined as
\begin{equation}
    T =
    \begin{pmatrix}
        \R & \mathbf{X} & \mathbf{V} \\
        0_{1,3} & 1 & 0 \\
        0_{1,3} & 0 & 1 \\
    \end{pmatrix}
\end{equation}
where $0_{1,3}$ is a 3 dimensional zero row vector.

For our metric, we need to define the group composition and inverse actions. The group composition action is defined as
\begin{equation}
    T_1 * T_2
    =     \begin{pmatrix}
        \mathbf{R_1R_2} & \mathbf{R_1X_2 + X_1} & \mathbf{R_1V_2 + V_1} \\
        0_{1,3} & 1 & 0 \\
        0_{1,3} & 0 & 1 \\
    \end{pmatrix}
\end{equation}
and the inverse operation is defined as
\begin{equation}
    T^{-1}
    =     \begin{pmatrix}
        \mathbf{R^{-1}} & \mathbf{-R^{-1}X} & \mathbf{-R^{-1}V} \\
        0_{1,3} & 1 & 0 \\
        0_{1,3} & 0 & 1 \\
    \end{pmatrix}
\end{equation}
The definition of $\SEtwothree$ thus aids us in defining closed form group operations which are also efficient to compute.

While we don't explicitly use the Lie algebra aspects of the $\SEtwothree$ group in our metric, these properties are beneficial for other aspects, such as alignment as discussed below.

\subsection{Alignment}
The trajectories first need to aligned since they can have various ambiguities and gauge freedoms, such as being in different coordinate frames. The transformation required to perform the trajectory alignment is denoted by $\textbf{S}$.
To handle these ambiguities, we follow the alignment process detailed in~\citet{Zhang18iros} since it provides formulae for each component of our generalized state vector.

Let the similarity transformation be parameterized by $S = \{s, \R, \mathbf{t}\}$, where $s \in \mathbb{R}$, $\R \in \SOthree$, and $\mathbf{t} \in \mathbb{R}^3$. $S$ performs the alignment transformation as
$$
\R'_i = \R \R_i,~\pos'_i = s\R\pos_i + \mathbf{t},~\vel'_i = s\R\vel_i
$$

An additional benefit of the Lie group formulation of our approach is that the trajectory alignment can be performed as optimization on the manifold. By leveraging the Lie algebra, we can extend the exposition of~\citet{Salas15rss} for the optimal alignment, while taking into consideration all components of the generalized state vector.

For the rest of our discussion, we assume ambiguity resolution and trajectory alignment has already been performed, and the alignment transformation is denoted by $\mathbf{S}$.

\subsection{Absolute State Error}

We denote our proposed metric as \textit{Absolute State Error} to represent the global nature of the metric as well as fact that it explicitly models state estimates. Given an estimated state trajectory $\mathbf{P}_{1:N} \in \SEtwothree$ and the ground truth states as $\mathbf{Q}_{1:N} \in \SEtwothree$, alongwith the alignment transformation $\mathbf{S}$ we can compute the Absolute State Error at time step \textit{i} as
\begin{equation}
    \mathbf{E}_i = \mathbf{Q}^{-1}_i \mathbf{S} \mathbf{P}_i
\end{equation}

To evaluate the state estimates over all time indices, we compute the root mean square error similar to~\cite{Grupp17evo}:
\begin{equation}
    \label{eqn:ase_rmse}
    \text{RMSE}(\mathbf{E}_{1:N}) = \Bigg( \frac{1}{N} \sum_{i=1}^{N} \Vert \mathbf{E}_i - \I_{5\times5} \Vert^2_F \Bigg)^{1/2}
\end{equation}
where $F$ is the Frobenius matrix norm.

The observant reader will notice that the error metric is of a form similar to the \textit{Absolute Trajectory Error} proposed by~\citet{Sturm12iros_rgbd_slam}. This is advantageous since existing software libraries and tools can be easily extended to use our metric for state estimation.

\section{Chebyshev Polynomial Interpolation}
In this section, we briefly review the Chebyshev series and interpolation using Chebyshev polynomials of the second kind (henceforth referred to as Chebyshev polynomials for brevity). We primarily follow the exposition in~\cite{Trefethen13book}, to which we refer the reader for more details on numerical analysis using Chebyshev polynomials for function approximation.

\subsection{Chebyshev Series \& Polynomials}
A Chebyshev series is an infinite series akin to the Fourier series which forms an orthogonal basis defined on a unit circle.
It is defined for functions $f(x)$ on the interval $\left[-1, 1\right]$, though it can be scaled to arbitrary bounds (\eg via min-max normalization), and the series decomposition is given by
\begin{equation}\label{eq:series}
    f(t)=\sum_{k=0}^{\infty}a_{k}T_{k}(t)\;\;\mbox{with}\;\;a_{k}=\frac{2}{\pi}\int_{-1}^{1}\frac{f(t)T_{k}(t)}{\sqrt{1-t^{2}}}dt
\end{equation}
with the factor $2/\pi$ changing to $1/\pi$ for $k=0$.
Above, $T_{k}$ is the $k$th Chebyshev polynomial\textbf{,} defined as the projection of a cosine function to the midline $\left[-1,1\right]$ of the unit circle. 
\begin{eqnarray}
    \label{eq:Tk}
    T_{k}(t) & \triangleq & \cos\left(k\arccos(t)\right), \; -1 \leq t \leq 1
\end{eqnarray}


Given an arbitrary real function $f$ on $\left[-1,1\right]$, we can find the best approximation to $f$ with a Chebyshev polynomial of degree $N$. This is due to the Weierstrass Approximation Theorem~\cite{Trefethen13book}) which we recapitulate below

\begin{theorem}[Weierstrass Approximation Theorem]
    Let $f$ be a continuous function on $\left[-1,1\right]$, and let $\epsilon>0$ be arbitrary.
    Then there exists a polynomial $p$ such that
        \begin{equation}
            \Vert f - p \Vert < \epsilon
        \end{equation}
\end{theorem}
A generalized proof can be found in~\cite{Stone48mathmag}.
Thus, we can find the best approximation to any function by interpolating at the Chebyshev-Gauss-Lobatto (or simply Chebyshev) points, defined as:
\begin{equation}
    t_{j}\triangleq\cos\left(j\pi/\degree\right),~0\leq j\leq\degree\label{eq:chebpoints}
\end{equation}

\begin{figure}[h!]
    \captionsetup{font=footnotesize}
    \centering{}\includegraphics[viewport=70bp 280bp 550bp 515bp,clip,width=0.7\columnwidth]{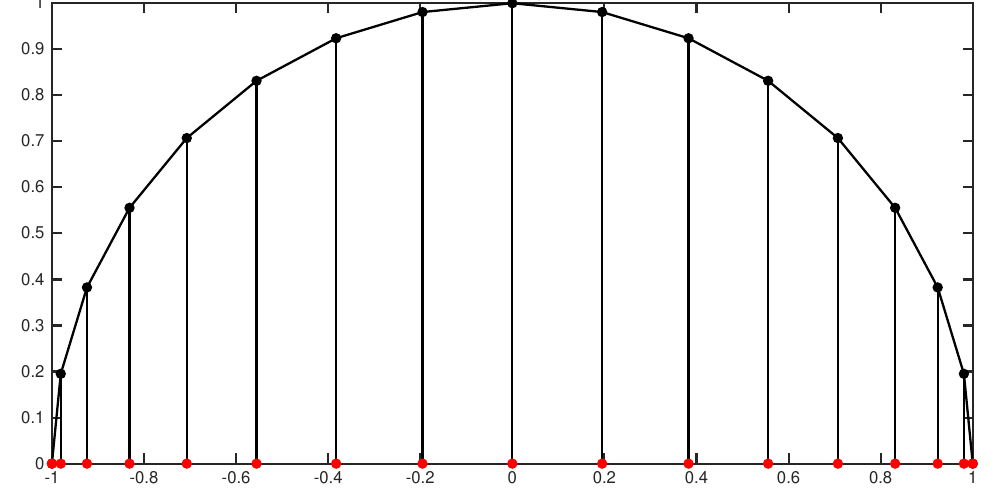}
    \caption{Chebyshev points $\cos(k\pi/n)$ for degree $n=16.$ They are obtained by projecting a regular grid on the unit circle onto the x-axis.}
    \label{fig:Chebyshev-Gauss-Lobatto}
\end{figure}

Interpolation at Chebyshev points ensures efficient convergence of the approximation and avoids issues such as Runge's phenomenon~\cite{Berrut04siam} which causes undesirable oscillation of the interpolant at the endpoints.
Furthermore, given the interpolant values, we can make use of the Fast Fourier Transform (FFT) to efficiently obtain the coefficients $a_{k}$ of the truncated Chebyshev series (the Chebyshev series upto a specified degree $N$). A visualization is provided in figure~\ref{fig:Chebyshev-Gauss-Lobatto}.

\subsection{Barycentric Interpolation}

Given the samples of $f$ at the $\degree+1$ Chebyshev points $f_j$, we can evaluate any point in $f$ efficiently using the Barycentric Interpolation formula~\cite{Berrut04siam}.
Computationally, the Chebyshev points provide an advantage over equidistant sampled points due to the simplicity of the resulting Barycentric weights, giving us the following interpolation formula
\begin{equation}
    \label{eq:barycentric}
    f(x) = \sum_{j=0}^{\degree}\frac{(-1)^jf_j}{x-x_j} \Big/ \sum_{j=0}^{\degree}\frac{(-1)^j}{x-x_j} 
\end{equation}
with $f(x)=f_j$ if $x=x_j$, and the summation terms for $j=0$ and $j=\degree$ being multiplied by $1/2$.

An additional benefit of Barycentric Interpolation is that it yields a linear form which can be computed via matrix-vector products. Since this is a linear operation, we can reparameterize (\ref{eq:barycentric}) as an efficient inner product
$ f(x) = \textbf{f} \cdot \textbf{w} $
where $\textbf{f}$ is a vector of all the values of $f$ at the Chebyshev points, and $\textbf{w}$ is an $(\degree+1)$ vector of Barycentric weights.

\subsection{Differentiation via Interpolation}
\label{sec:diff-via-int}
While finite differences is a common approach for computing derivatives, it is superseded by \textit{Differentiation via Interpolation}. Here, an interpolating polynomial is fit to the samples from the function $f(x)$ (in the least-squares sense) and the derivative of the interpolant is used for computing $f'(x)$. In general, it can be shown that finite differences is a special case of differentiation via interpolation~\cite{Driscoll17book}.
Furthermore, since polynomial interpolation is a global method, it has better precision and smoothing properties compared to local methods such as (centered) finite differences~\cite{Ahnert07compphyscommun}.

Spectral collocation methods, commonly used in the study of differential equations~\cite{Hussaini89anm}, yield an efficient process for obtaining derivatives of the approximating polynomial via the differentiation matrix.
Thus, the derivatives of the interpolating polynomial can be efficiently computed via a matrix-vector product. This is useful when performing optimization as we can compute the derivatives of arbitrary functions for (almost) free.

\subsection{Fitting to Noisy Data}
\label{sec:fit-data}
\global\long\def\weights{\mathbf{w}}
\global\long\def\wi{\mathbf{\weights_{i}}}
\global\long\def\deltaX{\mathbf{\Delta}}
\global\long\def\state{\mathbf{x}}
\global\long\def\states{\mathbf{X}}
\global\long\def\DN{\mathbf{D}_{\degree}}

The key idea is to express the least-squares polynomial fit in terms of the pseudo-spectral parameterization. Pseudo-spectral parameterization is a way of parameterizing a polynomial completely by its values at specific points, instead of with polynomial coefficients. Given the $m$ vector data samples $\mathbf{Z}$ at discrete time indices, we can minimize the following objective to get optimal the polynomial fit $p^*(\states)$ at the pseudo-spectral points:
\begin{equation}
    \label{eqn:objective-func}
    p^*(\states) = \arg\min_\weights \sum_{i = 1}^m \Vert \mathbf{z}_i - \states \cdot \weights_i \Vert_{\Omega}^2
\end{equation}
The main mechanism used is the barycentric interpolation formula~\ref{eq:barycentric} which predicts the state $\state(i)$ at any arbitrary time $i$ as $\state_i = \states \cdot \weights_i$, with $\Omega$ refering to the covariance of the measuring device. As a least-squares objective~\ref{eqn:objective-func} can be minimized using non-linear programming techniques.

\section{COMPUTING TRUE LINEAR VELOCITY}

In order to evaluate state estimation algorithms, it is important to collect ground truth pose and linear velocity data. While motion capture devices and laser trackers are often used for collecting pose data, linear velocity data is not as straightforward to obtain.

We propose to use differentiation via interpolation to estimate the ground truth linear velocity.
While most approaches compute the linear velocity using the method of finite differences, differentiation via interpolation is theoretically advantageous, as discussed in section~\ref{sec:diff-via-int}. Given the ground truth translation data from the motion capture device, we use this data as our noisy measurements in~\ref{eqn:objective-func}. By performing a nonlinear least squares optimization, we can obtain a pseudo-spectral parameterization $\mathbf{C}$ of an interpolating polynomial of degree $\degree$ which represents the translation data.
We can use the covariance of the motion capture device as the values for $\omega$ to specify our belief in the device's accuracy.

Finally, to compute the true linear velocity, we multiply the values of the polynomial with the $(\degree+1)\times(\degree+1)$ differentiation matrix $\DN$, as defined in \cite[p. 53]{Trefethen00book}:
\begin{equation}
    \label{eqn:velocity-computation}
    \vel = \dot{\pos} = \DN \mathbf{C} \weights_{i}
\end{equation}
\section{EXPERIMENTAL RESULTS}

We experimentally validate our Chebyshev interpolation scheme and the use of our proposed C-ASE metric.
For the interpolation scheme, we show results in simulation on autonomous driving and quadruped walking.

\begin{figure}[h!]
    \captionsetup{font=footnotesize}
    \centering
    \begin{subfigure}{0.4\textwidth}
        \includegraphics[width=\textwidth,height=0.55\textwidth]{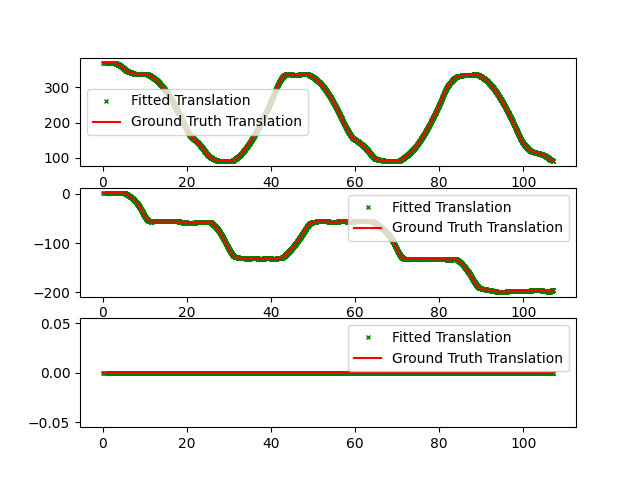}
    \end{subfigure}
    \begin{subfigure}{0.4\textwidth}
        \includegraphics[width=\textwidth]{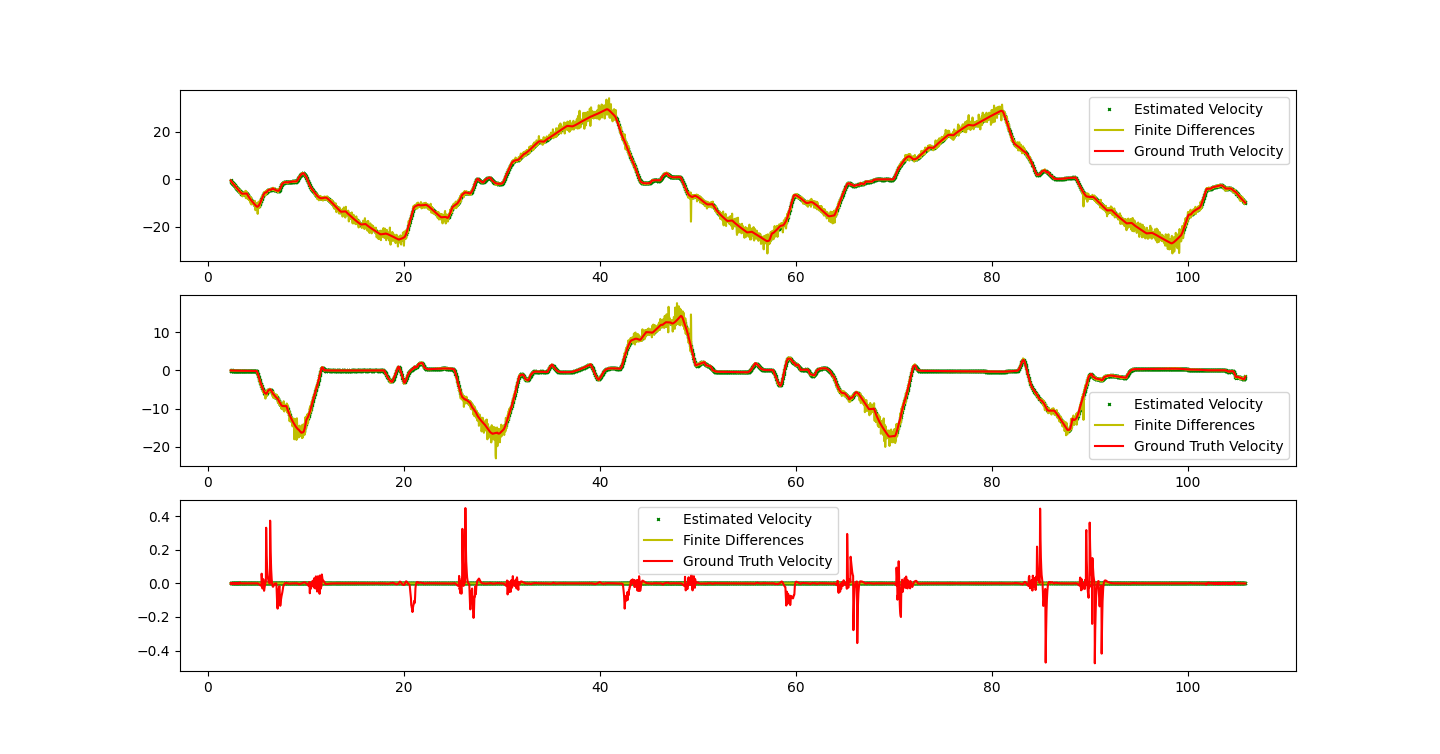}
    \end{subfigure}
    \caption{Chebyshev polynomial fit to the trajectory collected from the CARLA simulator. The ground truth translation and linear velocities are shown as red lines, and the fitted translation and computed velocity are shown as green dots. Result from (centered) finite differences is also shown for comparison. The degree of the polynomial was arbitrarily chosen as 400.}
    \label{fig:carla}
    \begin{subfigure}{0.5\textwidth}
        \includegraphics[width=\textwidth]{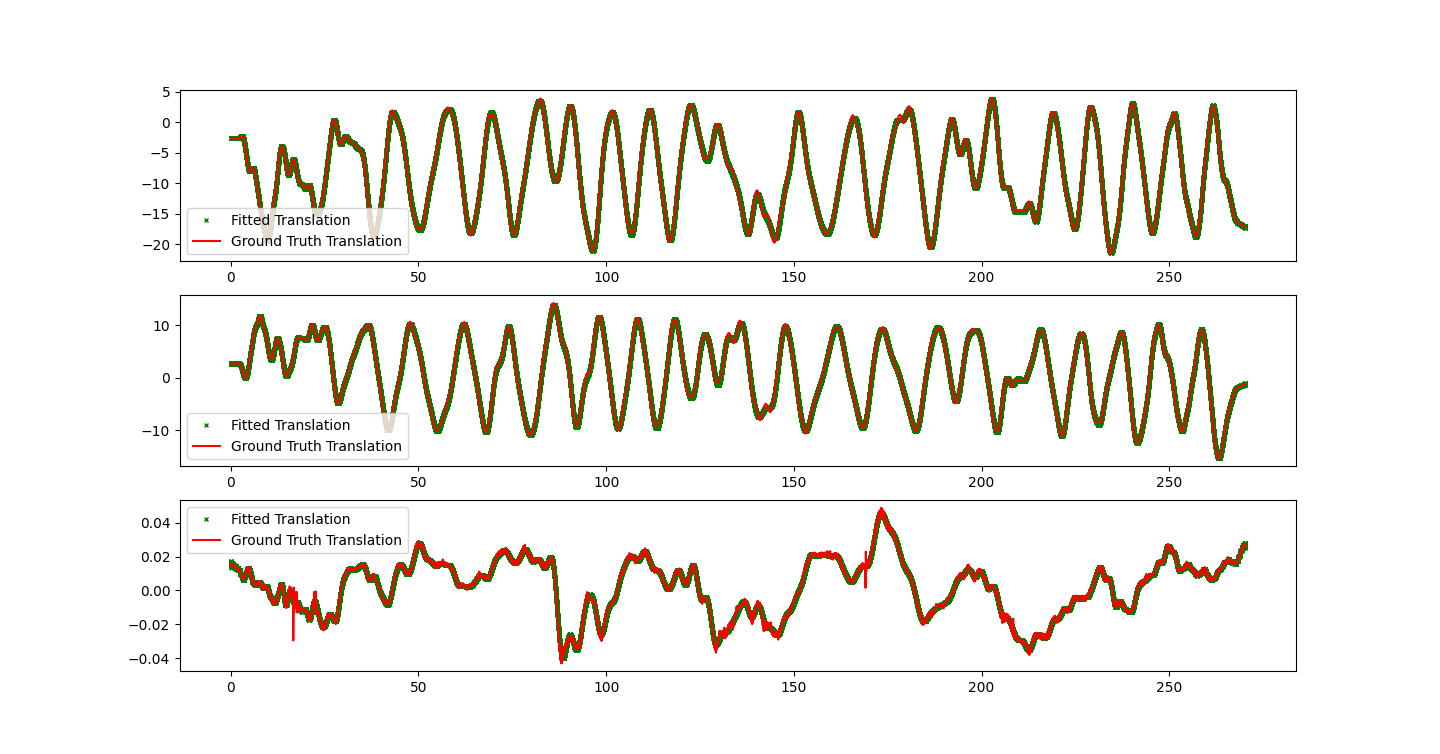}
    \end{subfigure}
    \begin{subfigure}{0.5\textwidth}
        \includegraphics[width=\textwidth]{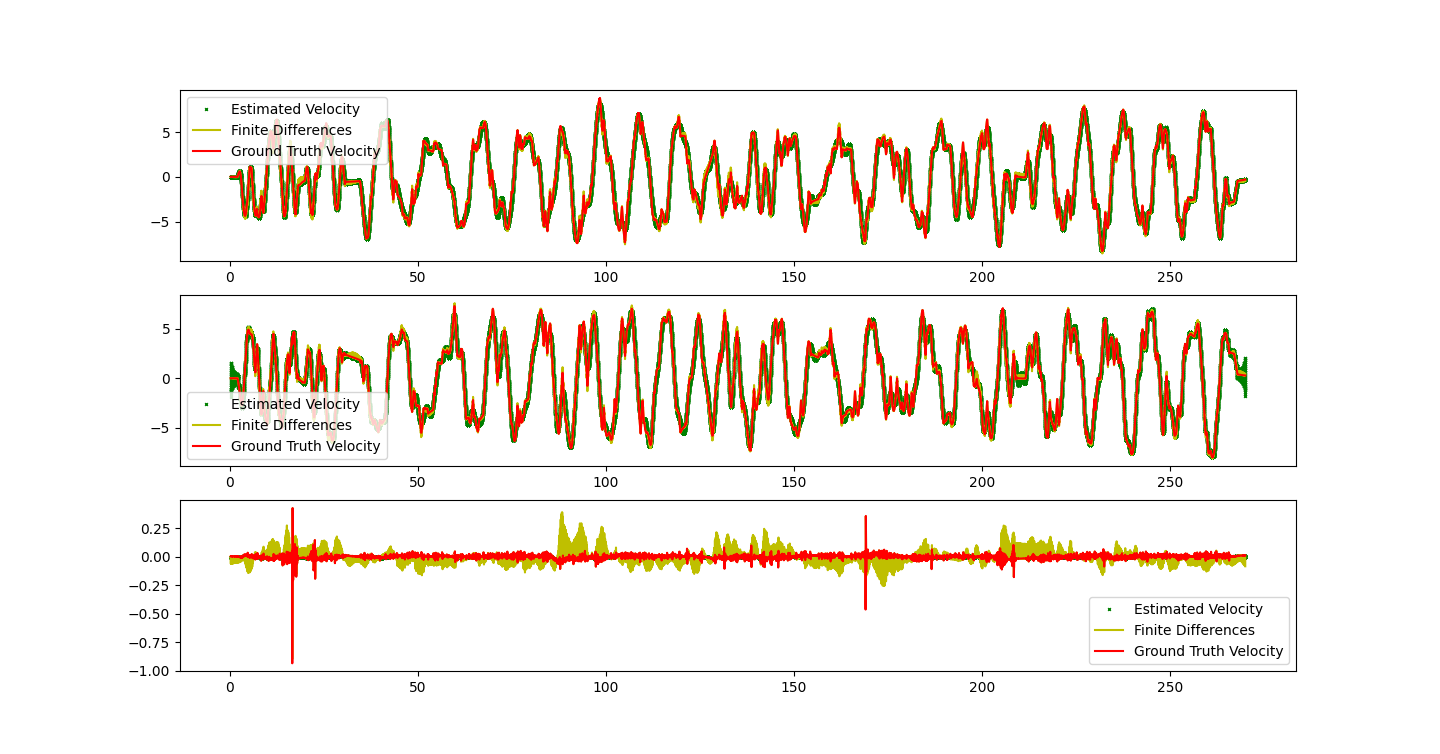}
    \end{subfigure}
    \caption{Similar to the CARLA simulator, we qualitatively show the results of Chebyshev interpolation and differentiation on the AutoRally data.}
    \label{fig:autorally}
\end{figure}

\begin{figure}[h]
    \captionsetup{font=footnotesize}
    \centering
    \begin{subfigure}{0.23\textwidth}
        \includegraphics[width=\textwidth]{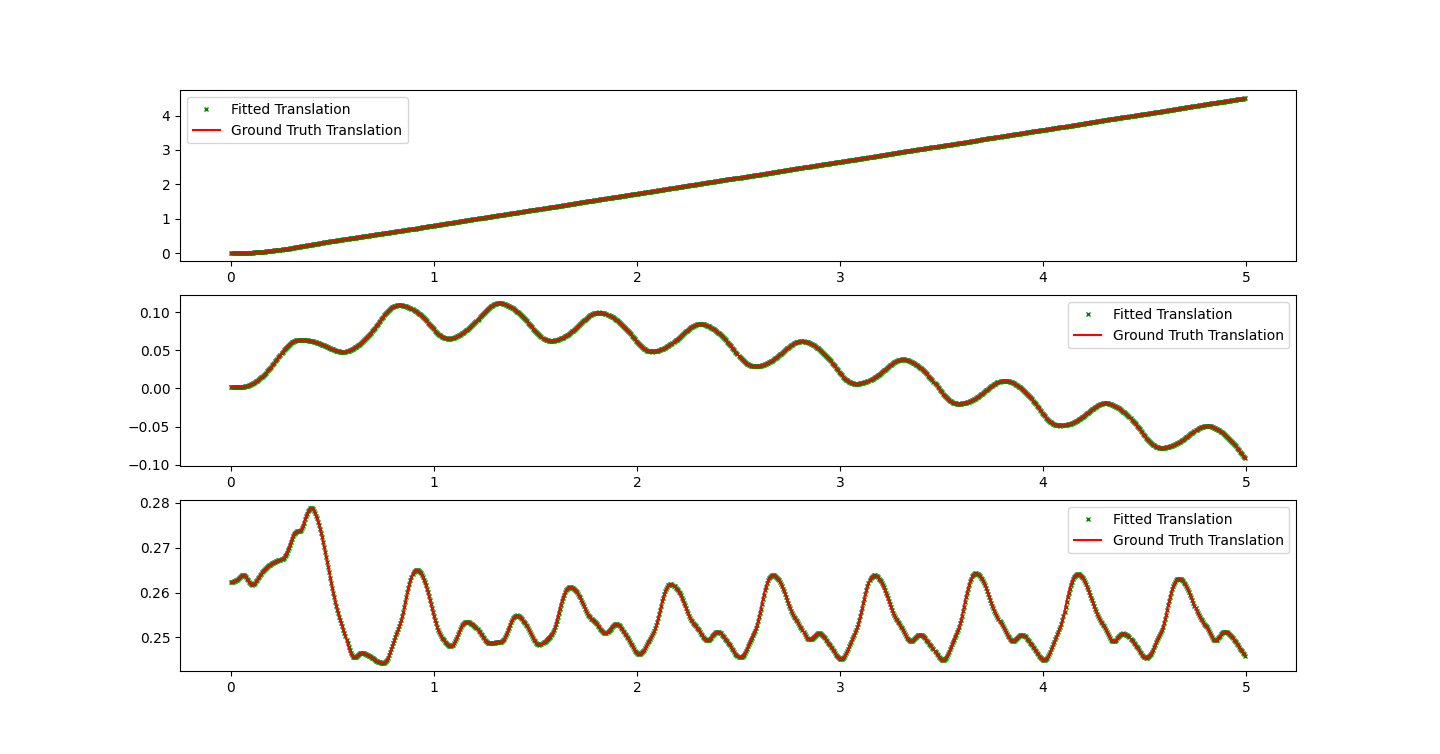}
    \end{subfigure}
    \begin{subfigure}{0.24\textwidth}
        \includegraphics[width=\textwidth]{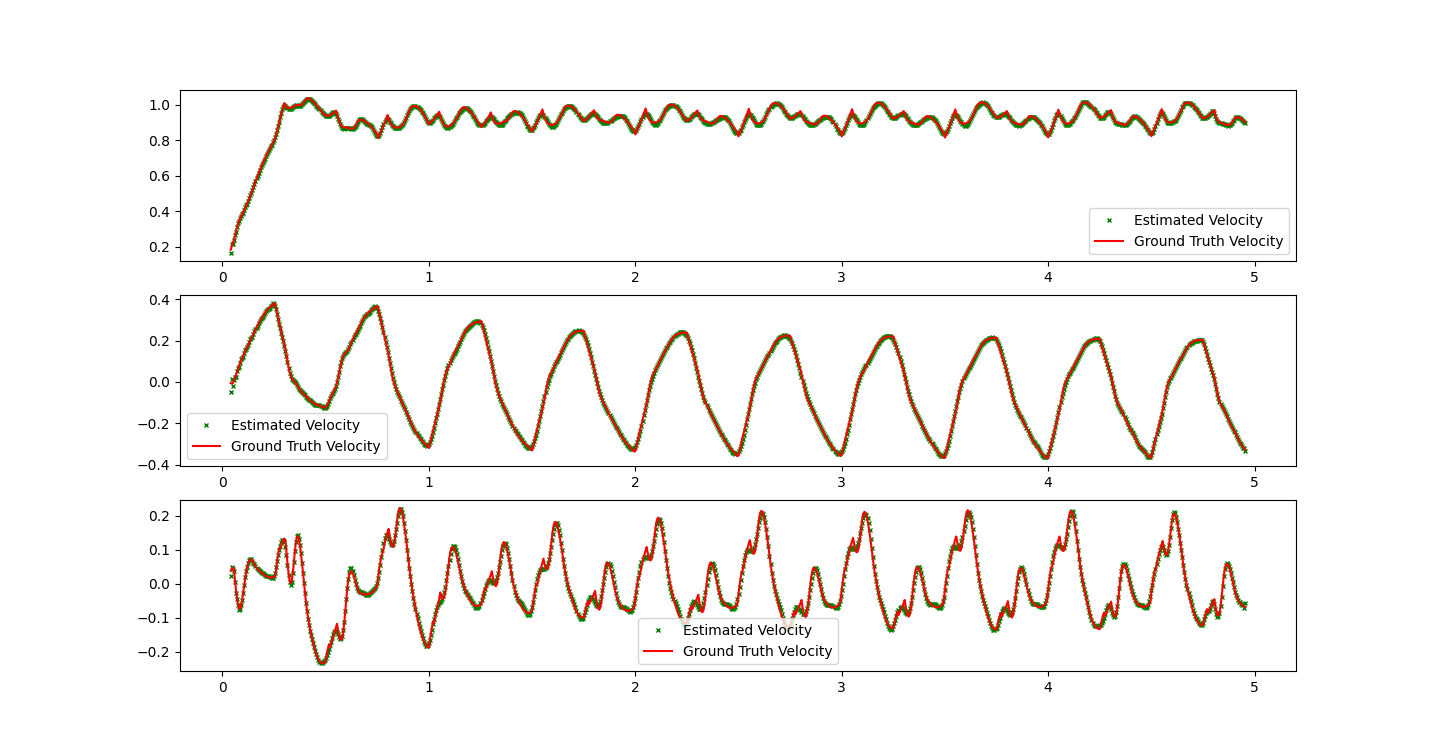}
    \end{subfigure}
    \begin{subfigure}{0.23\textwidth}
        \includegraphics[width=\textwidth]{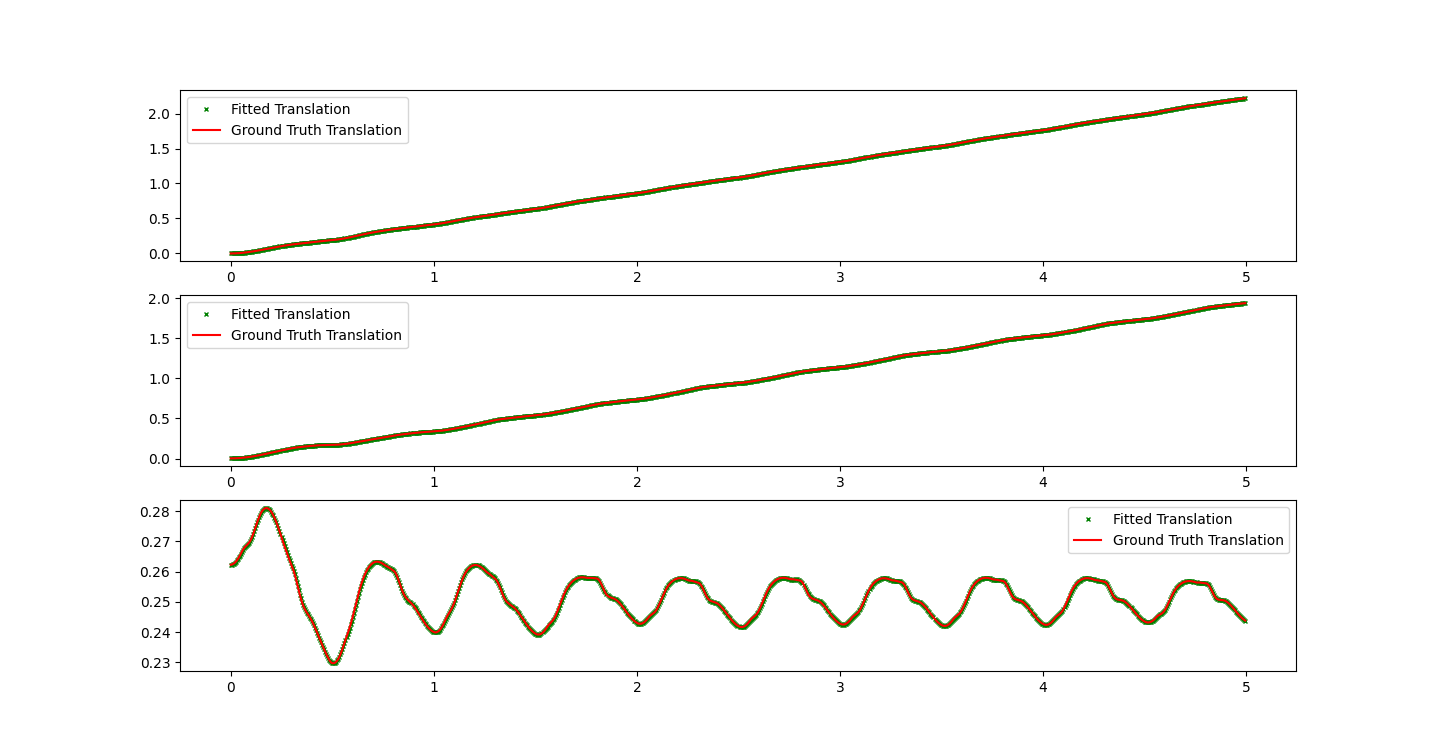}
    \end{subfigure}
    \begin{subfigure}{0.24\textwidth}
        \includegraphics[width=\textwidth]{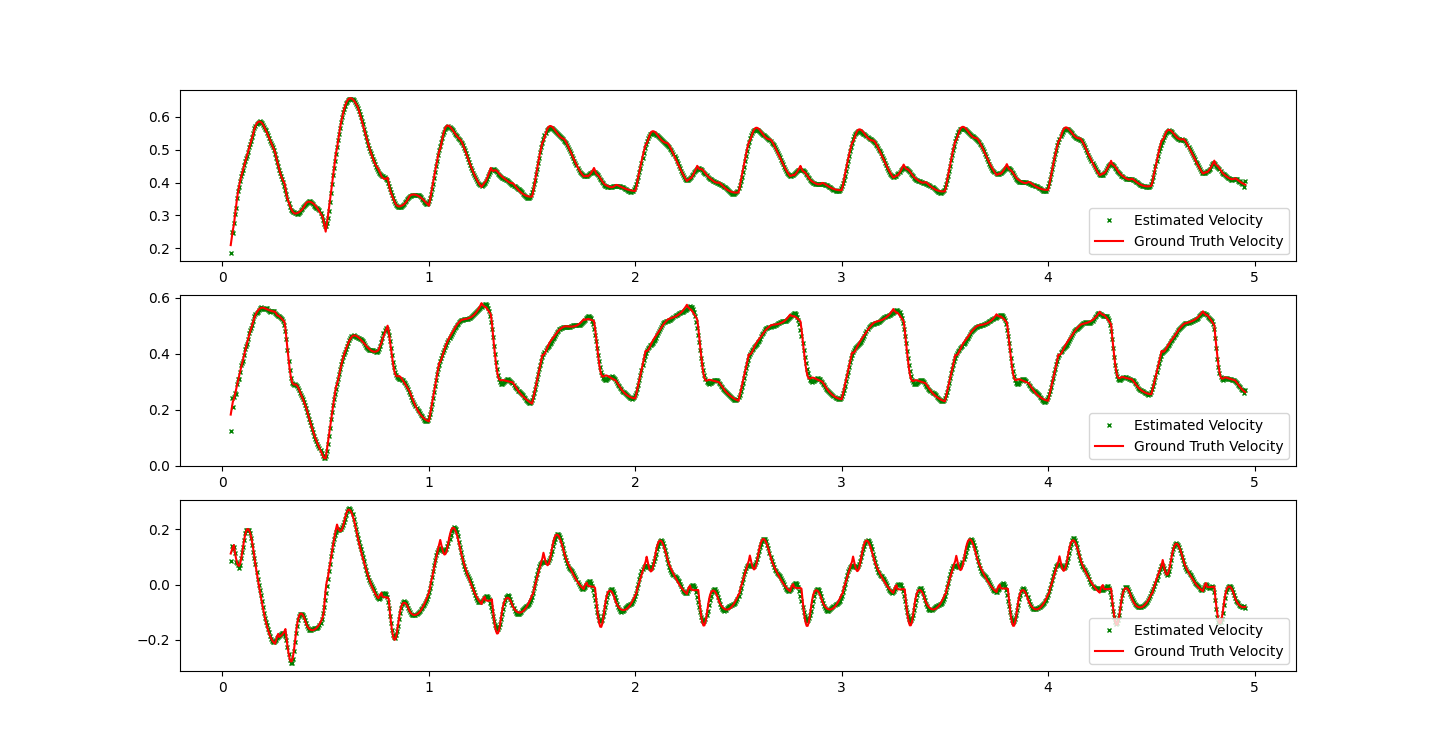}
    \end{subfigure}
    \begin{subfigure}{0.23\textwidth}
        \includegraphics[width=\textwidth,height=0.55\textwidth]{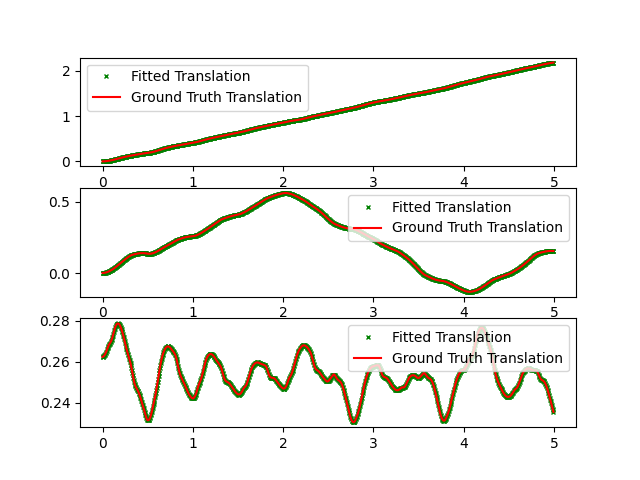}
    \end{subfigure}
    \begin{subfigure}{0.24\textwidth}
        \includegraphics[width=\textwidth]{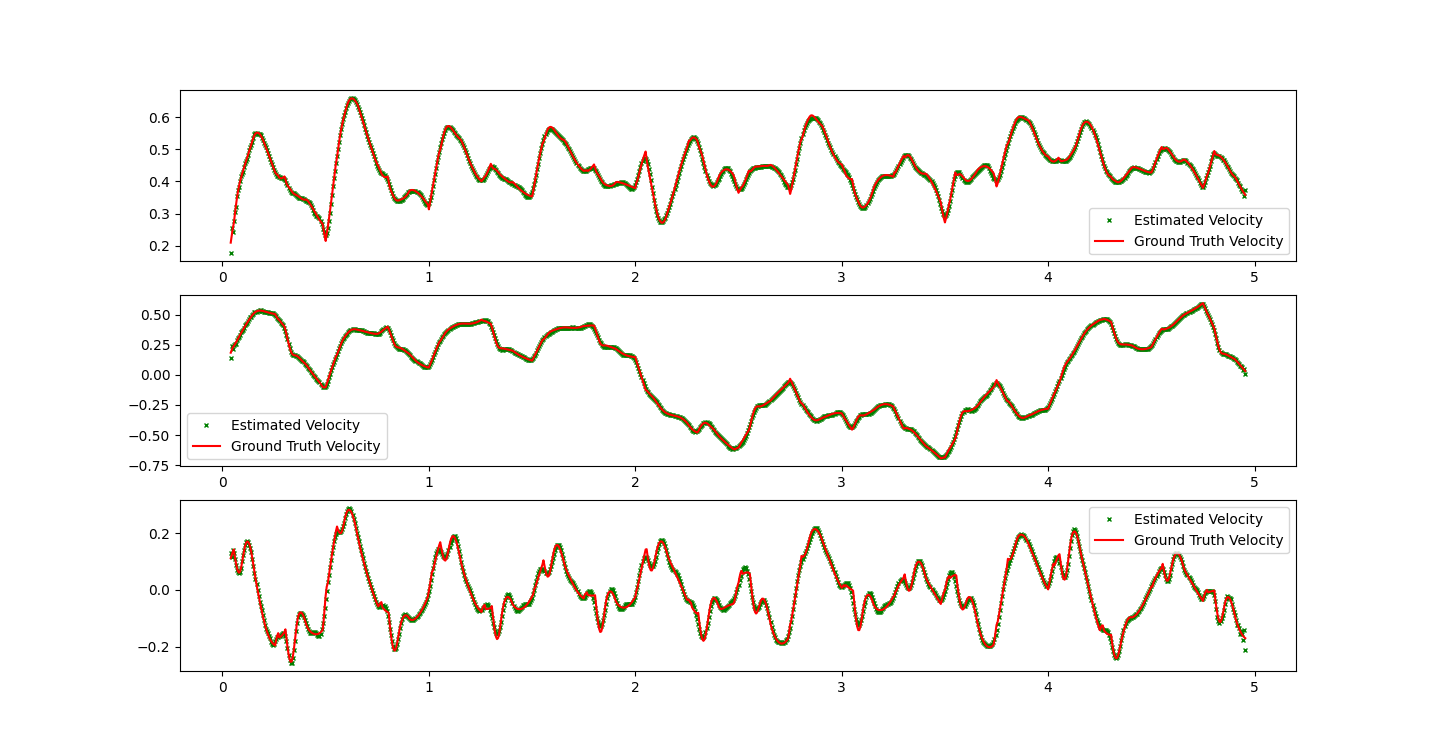}
    \end{subfigure}
    \caption{Chebyshev polynomial fit to the translation and velocity data of an A1 quadruped walking in a diagonal left-front trajectory. The ground truth data is collected in the Pybullet simulation environment~\cite{Coumans16report}.}
    \label{fig:a1-walking}
\end{figure}

\subsection{Autonomous Driving}

In the case of autonomous driving, we collect trajectories from two different simulators, the CARLA simulator~\cite{Dosovitskiy17corl}, and a simulated version of the AutoRally project~\cite{Goldfain19csm}. This allows us to have access to both ground truth translation and linear velocity data.
We collected a trajectory from each simulator by having a human control the vehicle for an arbitrary amount of time. By letting the trajectory vary in length and possible states, we wish to show the generality and accuracy of our polynomial interpolation approach.

Figs.~\ref{fig:carla} and \ref{fig:autorally} show the qualitative results of Chebyshev polynomial interpolation on the translation data. For each trajectory's position data, we correspondingly compute the linear velocity as per section~\ref{sec:diff-via-int} and plot that against the true velocity along with the result of using centered finite diferences. From the plots, we can see that the differentiation via interpolation scheme performs at least as well as finite differences, while also possessing a \textit{smoothing} effect on the velocity data, \ie it avoid extreme peaks.

For quanitative comparison, we compute the Root Mean Square Error (RMSE) between the fitted translation and the true translation, as well as the computed trajectory and the ground truth trajectory. This allows us to examine the effectiveness of our approach at representing the data, as well as how well differentiation via interpolation can perform. For the CARLA trajectory, the RMSE between the polynomial-computed velocity and the true velocity is $0.1860$ compared to the RMSE when using finite differences $1.2025$. We see a similar trend for the AutoRally trajectory. Table~\ref{table:polynomial-fit-results} provides a summary of the experimental results.

It is important to note that the above interpolation was performed with a view of demonstrating correctness, rather than finding the optimal polynomial degree. A higher degree polynomial would lead to lower RMSE, and due to the pseudospectral nature, would still provide significant data compression in terms of representation compared to storing the trajectory in its raw format.

\subsection{Legged Locomotion}

To further demonstrate the generality of our approach, we apply the differentiation via interpolation scheme to trajectories of a walking quadruped robot. We consider 3 trajectories: an A1 robot walking in a straight line, in a diagonal line, and in a zig-zag motion. From fig.~\ref{fig:a1-walking} we see similar results to the autonomous driving experiments, where the translation data is fit accurately, and the computed linear velocity has low RMSE compared to the ground truth. The quantitative summary of the results is provided in table~\ref{table:polynomial-fit-results}.

\begin{table}[h]
    \centering
    \captionsetup{font=footnotesize}
    \caption{Experimental Results from the CARLA simulator and the AutoRally simulator for autonomous driving, and the A1 robot in the Pybullet simulator for quadruped walking.}
    \begin{tabular}{|c|c|c|c|c|}
        \hline
        \multicolumn{1}{|c|}{Trajectory} & Time (s) & {Transl. RMSE} & Vel. RMSE & Degree \\ \hline \hline
        \multicolumn{1}{|c|}{CARLA}                   & 107.141 & 0.0438 & 0.1860    & 400 \\ \hline
        \multicolumn{1}{|c|}{AutoRally Sim}       & 270.23 & 0.3651      & 1.0586 &  256 \\ \hline
        \multicolumn{1}{|c|}{A1 Straight}            &  5.0 & 9.3101e-05    & 0.0121 & 200 \\ \hline
        \multicolumn{1}{|c|}{A1 Diagonal}           &  5.0 & 8.6859e-05  & 0.0114 & 200 \\ \hline
        \multicolumn{1}{|c|}{A1 Zig-Zag}            &  5.0 & 8.3434e-05  & 0.0111  & 200 \\ \hline
    \end{tabular}
    \label{table:polynomial-fit-results}
\end{table}

\subsection{Comparative Analysis}

Finally, we test our metric using two state estimators on robot hardware over a 50 second trajectory. For a fair comparison, we use the Bloesch state estimator~\cite{Bloesch12rss_se_imu_leg_kinematics} and the Two State Implicit Filter (TSIF) estimator~\cite{Bloesch17ral_two_state}, both of which are developed by the same group, with the latter shown to have better performance. This is used to verify whether our metric accurately captures the performance gap between these two filtering based state estimators when run on the Anybotics ANYmal C quadruped~\cite{Hutter16iros_anymal}. Ground truth pose data is collected with the use of motion capture, and the polynomial degree is $128$.

\begin{table}[h]
    \centering
    \captionsetup{font=footnotesize}
    \caption{Comparison of two filtering based state estimators from the same research group, demonstrating our C-ASE metric accurately captures the difference in performance. Note that the C-ASE metric is dimensionless.}
    \begin{tabular}{|c|c|c|c|}
        \hline
        \textbf{Estimator} & \textbf{RMSE} & \textbf{STD} & \textbf{Median} \\ \hline
        Bloesch Filter & 0.562489 & 0.226617 & 0.487763 \\ \hline
        TSIF       & 0.297052 & 0.152732  & 0.267481 \\ \hline
    \end{tabular}
    \label{table:ase-comparison}
\end{table}

As can be seen from table~\ref{table:ase-comparison}, the TSIF estimator has significantly lower C-ASE RMSE when compared to the Bloesch filter, reflecting that our metric indeed captures the difference in performance accurately.


\section{CONCLUSION}

In this work, we have proposed a new metric for robot state estimation evaluation, which leverages the properties of the recently proposed $\SEtwothree$ Lie group. Due to its group properties, this metric explicitly takes into account the linear velocity component of the generalized state vector, while providing a singular comparative value, and is easy to incorporate into existing software libraries and pipelines.

To facilitate the use of this metric, we also show how ground truth linear velocity can be accurately and efficiently computed from translation data obtained from motion capture devices. Leveraging polynomial interpolation using the Chebyshev series of the second kind and differentiation via interpolation, we experimentally validate the accuracy for multiple robotic platforms, and show better results compared to the more common centered finite differences method. Moreover, we leverage a pseudospectral parameterization which provides large data compression in terms of trajectory representation as an additional benefit.

We hope our proposed metric provides a unified approach to benchmarking state estimation systems, and encourages further research in approximation methods for robot trajectory representation.
This will allow a fair comparison of such systems, and hopefully  lead to further progress in the field of robotic state estimation.


\printbibliography

\end{document}